\documentclass[conference]{IEEEtran}
\IEEEoverridecommandlockouts
\usepackage{cite}
\usepackage{amsmath,amssymb,amsfonts}
\usepackage{algorithmic}
\usepackage{graphicx}
\usepackage{textcomp}
\usepackage{xcolor}
\usepackage[ruled,vlined]{algorithm2e}
\def\BibTeX{{\rm B\kern-.05em{\sc i\kern-.025em b}\kern-.08em
    T\kern-.1667em\lower.7ex\hbox{E}\kern-.125emX}}
\begin{document}

\title{Architecture for Multi-Unmanned Aerial Vehicles based Autonomous Precision Agriculture Systems}

\author{
\IEEEauthorblockN{
Ebasa Temesgen\IEEEauthorrefmark{1}\IEEEauthorrefmark{2},
Nathnael Minyelshowa\IEEEauthorrefmark{2},
Lebsework Negash\IEEEauthorrefmark{2}
}\\
\IEEEauthorblockA{\IEEEauthorrefmark{1}University of Minnesota}
\IEEEauthorblockA{\IEEEauthorrefmark{2}Addis Ababa Institute of Technology (AAiT)}
}

\maketitle

\begin{abstract}
%\boldmath
The use of unmanned aerial vehicles (UAVs) in precision agriculture has seen a huge increase recently. As such, 
systems that aim to apply various algorithms on the field need a 
structured framework of abstractions. This paper defines the various 
tasks of the UAVs in precision agriculture and model them into  an architectural framework. The presented architecture is built on the 
context that there will be minimal physical intervention to do the tasks 
defined with multiple coordinated and cooperative UAVs. Various tasks such as image processing, path planning, communication, data acquisition, and field mapping are employed in the architecture to provide an efficient system. Besides, different limitation for 
applying Multi-UAVs in precision agriculture has been considered in 
designing the architecture. The architecture provides an autonomous 
end-to-end solution, starting from mission planning, data acquisition 
and image processing framework that is highly efficient and can 
enable farmers to comprehensively deploy UAVs onto their lands. 
Simulation and field tests shows that the architecture offers a number 
of advantages that include fault-tolerance, robustness, developer and user-friendliness
\end{abstract}


% \fi
\IEEEpeerreviewmaketitle

\section{Introduction}

In recent years, Unmanned Aerial Vehicles(UAVs) are gaining considerable attention both in the research and commercial world. Especially the agricultural sector has seen such an interest in the research and development of UAVs that it owns up to a large portion of the commercial market for the UAVs. Today’s high-end UAV farming technology offers UAV-powered planting techniques that reduce planting costs by up to 85\% \cite{grouprefa}. Of course, this is because UAVs offer the chance to solve some of the challenges in the agriculture sector. Scouting land and crops, checking for weeds and spot treating plants, and monitoring overall crop health are among the areas that UAVs are widely used today. The world must keep innovating in agricultural UAV systems to ensure the sustainable production of food, as the planet is subjected to an ever-increasing population and man-made global warming problems that jeopardize the norm of food demand-to-supply proportion. 

Technology adoption of smart agriculture has already established paradigms to increase farm productivity and quality, as well as improving working conditions through the reduction of manual labor. Satellite-based remote sensing is one of the technologies integrated into agriculture and transforms labor-intensive agricultural practices to an automated level. Satellite imagery \cite{grouprefb} provides valuable insights into increased crop production, mitigating the risk of crop damage, and making farming more sustainable. However, the process of manufacturing, the cost of launching, and accessing these satellites are setbacks, especially for developing countries. Moreover, for the reasons that satellites have time cycles to gather information and low image resolutions as the sensing bounded by the cloud and dusts, making satellites imagery to be less effective platforms for remote sensing. 

UAVs deliver high-quality data frequently to provide insights into crop developments, identifying best practices accurately and affordably \cite{grouprefa1}. Integration of UAVs to agricultural farming is one of the recent emerging technological advances towards precision farming. It is based on remote sensing technology that includes data acquisition, transmission, processing, and storage systems \cite{negash2019emerging}.

% To simplify the use of UAVs, reducing its cost of operation, as well as increasing its safety, a high level of automation is desired. Multiple challenges limit the usage of UAVs in a large field of agricultural land for high-demand applications such as surveying and monitoring. 
The application of UAVs in a large field of agricultural lands such as surveying and monitoring poses several challenges. These are the battery life of the UAVs, range of communication, loads and additional packages on the UAVs, data transferring and analysis in real-time, safety, and fault tolerance among others.

Although a large number of research groups \cite{grouprefc,grouprefd,grouprefe,groupref6} are working to solve different aspects of these problems, only a few of them aim to develop a complete end-to-end thorough solution for full autonomy Multiple Unmanned Areal Vehicles system (Multi-UAVs) that combines different components, such as data acquisition, communication, and processing into a single architecture.

Multiple commercial and open-projects exist that aim to develop complete software architectures for UAVs in general \cite{groupref6a}. The Aerostack, PX4 Flight Stack, and the APM Flight Stack and so many more have produced an architecture with a good level of autonomy, but their architecture is limited to a UAV’s control and autonomy rather than the application they serve \cite{groupref6b}. Besides, the use of Mult-UAVs in agricultural contexts has been limited to a few studies.

This paper aimed to design a thorough and complete architecture and framework for UAVs to be efficiently used in agricultural farms. The framework provides a modular architectural organization to support fully autonomous flights and a versatile software framework to carry out a mission efficiently. Also, a software framework is included to receives a mission from the user and optimizes the path, using the algorithm, then transmit to the UAVs with the set of commands from the user, then controls the UAVs flight without human intervention, which then transmits back the data, which will be fed to the server using a nearby Wi-Fi connection, which will be then analyzed using the AI engine in the backend, to be finally organized to be presented for the user via an interactive dashboard.

The rest of the paper is Organized as follows: Section \ref{Arc} presents The Multi-UAVs Architecture in more detail. Then Section \ref{sof} will explain about the software framework. Section \ref{eval} will look at the both the software and hardware implantation and use case of the architecture. In the Finally section \ref{conc}, the paper concludes and present some points that can be add in future works.  

\section{Architecture}\label{Arc}
This Architecture aimed to break down tasks in Multi-UAVs design into different units, especially for Agricultural sector allowing engineers and developers to work on different problems. 

\subsection{High-Level Description}
\begin{figure*}[!t]
  \includegraphics[width=\textwidth]{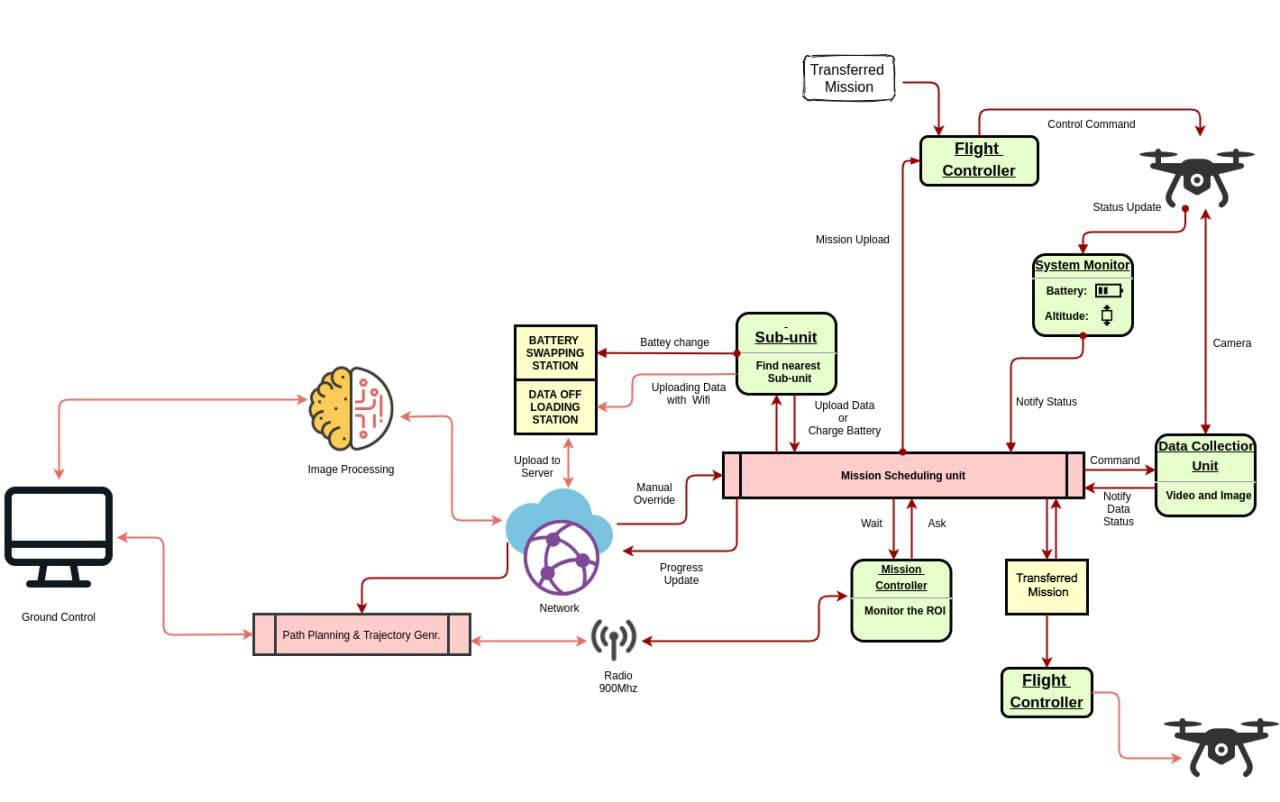}
  \caption{Multi-UAVs architecture, containing the communication scheme, different units and functionality starting from ground station control unit to the UAVs.}
  \label{ourarc}
\end{figure*}

In this architecture, fig. \ref{ourarc} both the centralized and decentralized approach \cite{groupref7} are used in decision-making, to enable each member  UAVs to be self-aware and to cooperate in executing a given task of monitoring and collecting data that is needed for analysis. The mission will be generated from the ground station, and loaded to all member UAVs, using the centralized approach, whereas control of execution and monitoring different factors will be decentralized, allowing each UAVs to decide on their own.  The presented design includes a ground station to specify the region of interest for all UAVs, path planning, trajectory generation, image processing unit on the deep learning algorithm, data offloading, and optimized battery swapping stations as shown in fig. \ref{ourarc}.

The centralized approach will allow the control and communication of each UAV to be centralized in the ground station, which has the advantage to ease in preparing the waypoints in the generated form region of interest and allocating path and trajectory for each UAV.  It's important to ensure that the path generated should have an optimal number of turns,  as it's been proved that more turns will lead to more battery usage. Also, areas assigned for each UVAs should be specific to avoid an overlap of areas to be covered.  This approach helps to reduce the complication that is caused in designing a mission for each UAVs and helps to reduce the communication and network traffic design. It will be also helpful to  see  the progress that recollection of data from the member UAVs to reconstruct a heat map.

The decentralized approach helps the UAVs to make decisions on their own, enabling them to self-monitor and creating situational awareness.  It gives them the ability to respond to problems (battery level, faults, etc.) that might arise when they are carrying out missions. It also provides the multi-agent systems to have coordination while carrying out a given task. During this time, one agent will be able to communicate with the other and be able to transfer its mission for the one capable of carrying it out. In the following section, will take a deeper look at some of the units in the architecture. 
\subsection{Communication}
Communication is one of the critical parts of any UAV Architecture.  Since most farmlands are flat and don't have an obstacles, the presented framework won’t be needing the design of new network architecture. Commercially available open-source hardware and software could used, with out a need for completion network design. This architecture used two types of network communications, namely radio communication and WiFi. The communications between the UAV and ground station and the data communication between the on-board computer to the server. The need to have a separate communication network is to avoid possible latency that will be caused in one unified network and to leverage the functionality of a flight of flight stack. This also provides flexibility where it can fit into any flight stack available. 

The radio communication such 3DR telemetry radio has a longer range and gives the ability to get data from the flight controllers such as PX4 or APM flight stacks and hence will be used as a communication module between the Ground station and the member UAVs. This will allow developers to use the rich communication protocol called  MAVLink \cite{groupref9} which is an acronym for Micro Air Vehicle Link. It follows a hybrid publish-subscribe, which is in multi-cast mode, that no additional overhead is generated and multiple subscribers can all receive this data and point-to-point design pattern, where MAVLink uses target ID and target component to communicate with the flight stack. Data streams are sent/published as topics while configuration sub-protocols such as the mission protocol or parameter protocol are point-to-point with re-transmission. 

After finishing the assigned task the UAVs will relocate themselves in the data offloading units where they upload the images and other data they gathered during their flight time and upload it from their onboard computer, which served as a temporary storage unit, to the server for analysis. The need to store the data and upload it through WiFi comes from the fact that streaming data directly is prone to a packet loss which re-transmission is not possible, especially during the flight of the UAVs. The socket server over the wifi network ensures a faster transmission rate.  
\subsection{Mission Generation}
A  mission generated by the system  will  always  be  planned  so  that  data  collection  or delivery can cover all parts of a land polygon. This is called Coverage Path Planning (CPP). For example, if a camera can take images in a 2 meters wide rectangle from some height, its possible sweep the land in 2 meter wide distances and collect data of each piece of land. Some constraints need to be considered before deploying a CPP system. For one, the energy usage of the UAV throughout its mission must be analyzed as it influences the area a UAV can cover with one charge which in turn affects the number of drones one must deploy to cover the whole area. As various papers have shown energy consumption can also be affected by the weight of the drone and environmental conditions like wind\cite{groupref10}.

In their paper, Ghaddar and Merei\cite{groupref13} use a trapezoidal tractor mobility path pattern to guarantee CPP. Other proposed methods also guarantee CPP\cite{groupref14,groupref15}. These methods in addition to the one proposed by Ghaddar and Merei have been weighted based on their performance in the length of the path they generate and the number of turns in the path (minimizing both will lead to better performance). The method that they proposed decreased the route length, the number of turns, the sum of the degrees in all turns, and more importantly completion time and energy consumption. This method also has an added advantage as it can easily be implemented and the path it generates can easily be split up to be fed to multiple UAVs. Thus, following factors are taken into account in mission generation. 

\begin{itemize}
    \item Important area priority: Prioritizing missions will be handy when there is scarcity in the amount UAVs that are needed for the given task. In this case, it's important to take input from the user to prioritize areas that the UAVs should cover. In this case, the user will select which regions have more priority over others. One way to do this is by giving an interface to set the distance between successive sweeps so that even if every point in the farmland polygon is not covered by the camera onboard the UAV.
\item The size of the Farm: As the farm area increases, it will be difficult to collect data especially if there are no sufficient number of UAVs available. Perhaps, one of the ways to avoid this is to take the UAVs to a higher altitude, which might bring a trade-off between the image qualities and other data resolution to a flight time. 
\end{itemize}

\subsection{Task Allocation}
In designing Multi-UAVs systems, it's crucial to have a well-designed system for decision making and a clear task allocation. In fact, this is one of the highly researched and challenging areas in the Multi-UAV system as it lays the foundation for coordination among each UAV. In farmlands, the task allocation becomes challenging as the shape of the farm gets complex and the size gets larger. Besides, it's important to make sure there are no over lapped region in mission generation and allocation in order to avoid the No-Fly Zones, so as to save the power from being wasted.  

In the presented architecture, the task allocation is based on the following factors.
\begin{itemize}

 \item Current Battery Level: UAV batteries have limited power capacity and it's important to consider them when assigning a task. Before a mission is assigned for each UAV the battery level they have will be first checked and compared with the threshold value. Depending on the comparison, either assign a task to the UAV or wait until the battery is more than the threshold. The architecture provide an interface so that a farm manager can set this threshold value manually with the minor caveat of setting a minimum threshold value to 10\%. 
 \item Flight History: As the UAVs get older, and frequently used, the power consumption, battery discharging rate, and performance will generally decrease. To compensate for this, the flight time and the battery level would have to be measured during the flight and it should be considered as input to estimate the flight time of a member UAV. The system will collect the amount of area covered and the battery drainage rate during the UAV’s time on air. This will help us determine what percentage of the mission every single UAV can carry out in successive trips. It also helps us estimate the battery lifetime, so that the farm manager can be alerted. 
\end{itemize}
Since the mission generated will be a set of points the UAVs will travel through (also the turning points), its possible to split it up based on the parameters stated above. This will be based on the following formula:
\begin{equation}
    mission = \{p_1, p_2, ... ,p_n \} 
\end{equation}
Its possible to segment this set based on distances estimated, which will be sustained by each UAV. This distance (d) can be calculated using: 
\begin{equation}
    d = \frac{ b - 10}{c}
    \label{eq1}
\end{equation} where b is the current battery level and c is a constant calculated after each successful mission as 
\begin{equation}
    c = \frac{  b_i - b_f}{l}
\end{equation}
where ${b_i}$ is the initial battery level and ${b_{f}}$ is the final battery level after the UAV lands. l is the distance that the UAV traveled in the previous mission. Its possible to segment the mission into n segments ${S_{m}}$ from where m goes from 0 to n where 
\begin{equation}
    S_m = \{p_i, p_{i+1}, ... , p_j\}
\end{equation}
with the constraint that 
\begin{equation}
 \sum_{k=i+1}^{j} dist(p_k,p_{k-1})  < d 
\end{equation}

where dist is a Euclidean distance function. 

\subsection{Battery swapping station}
Deployment of UAVs especially for monitoring and data collection that is carried out in large-scale farms requires multiple UAVs. One of the issues with Multi-UAV deployment is the fact that batters need to be reached to carry out missions rapidly since the lifetime of a single battery is low. Especially, without optimal charging positions, the UAV has to waste a valuable amount of energy returning to the charging station \cite{groupref8}. To avoid these complications, the architecture have incorporated battery swapping stations into the architecture.

When carrying out a mission, the UAV will be monitoring the battery level and the distance of the path left and the distance to the battery swapping station. Depending on the current battery level, it will estimate the probability of finishing the given task. If the chance of finishing the mission is higher then it will continue the mission, if not,  it will notify the ground station and will communicate with the nearest member to transfer its mission to the neighboring UAVs or cache for its next mission. That way all of the areas of interest will be covered through the usage of one or more of the UAVs.

% \begin{algorithm}
% \SetAlgoLined
% \KwIn{\textbf{begin}}
% %\KwResult{Write here the result }
% initialize \beta_{Threshold}, WayPoints,\\
%     get \beta_{Current}, P_{LAT}, P_{LONG}, BSS_{LAT},\\ BSS_{LONG},HOME_{LAT}, HOME_{LONG},
 
% \While{P_{LAT}, P_{LONG} \neq HOME_{LAT}, HOME_{LONG}}{  
%   Find a probability of finishing a mission using Eq.something \\
%      \eIf{\beta_{Current} > \beta_{Threshold}}{

%           \eIf{Probability > 0.5}{
%           WayPoints=+1\;
%           %instructions2\;
%           }{
%           Notify the GS\;
%           Set Up connection to nearest Drone\;
%           \eIf{Other UAV can finish the mission}{
%           transfer mission\;}
%           {cache it for later and return to BSS}
%           }
%      }
%      {
%      return to BSS
%      }
%  }
%  \caption{Our BSS algorithms}
% \end{algorithm}

\begin{algorithm}[t]
\SetAlgoNoLine
\SetAlgoNoEnd
\caption{Proposed BSS Algorithm}
\label{alg:bss}

Initialize $\beta_{\text{Threshold}}$ and $\mathrm{BSS}$\;
Get $P_{\text{Lat}}$ and $P_{\text{Long}}$\;

\While{$(P_{\text{Lat}}, P_{\text{Long}}) \neq \mathrm{BSS}$}{
    Get $\beta_{\text{Current}}$\;

    \eIf{$P_{\text{cached}} \neq \emptyset$}{
        $i \gets 0$\;
        \While{$i < \mathrm{length}(P_{\text{cached}})$}{
            Check whether $P_{\text{cached}}[i]$ is reachable using \eqref{eq1}\;

            \eIf{$\beta_{\text{Current}} > \beta_{\text{Threshold}}$}{
                $P \gets P_{\text{cached}}[i+1]$\;
            }{
                Notify the GS\;
                Set up connection to the nearest UAV\;
                \If{mission transfer is possible}{
                    Transfer mission\;
                }
                $P \gets \mathrm{BSS}$\;
            }

            Go to $P$\;
            Update $P_{\text{cached}}$\;
            $i \gets i + 1$\;
        }
    }{
        $i \gets 0$\;
        \While{$(P_{\text{Lat}}, P_{\text{Long}}) \neq \mathrm{BSS}$}{
            Check whether $P_{\text{waypoints}}[i]$ is reachable using \eqref{eq1}\;

            \eIf{$\beta_{\text{Current}} > \beta_{\text{Threshold}}$}{
                $P \gets P_{\text{waypoints}}[i+1]$\;
            }{
                Notify the GS\;
                Set up connection to the nearest UAV\;
                \If{mission transfer is possible}{
                    Transfer mission\;
                }
                $P \gets \mathrm{BSS}$\;
            }

            Go to $P$\;
            Update $P_{\text{cached}}$\;
            $i \gets i + 1$\;
        }
    }
}
\end{algorithm}
%\IEEEpubidadjcol

\section{Software Framework}\label{sof}
The presented architecture provides a software framework that allows end-users to hide the inner components of the system. The framework is also directly built upon the architecture and thus offers all advantages the architecture gives us in addition to being extended to add multiple functions in data processing or delivery of items. The main advantages of the software framework are:
\begin{itemize}
\item Scalability: the biggest advantage of the software framework and architecture is its scalability. The ease of adding additional drones and components into the system without change in the internal process enables a streamlined deployment process that saves time for farmland owners. Processes are also distributed over the architecture and this makes the system faster and more efficient.
\item Abstraction and User-friendliness: The software dashboard hides the detail about the functions in the process that will be hard to understand to farmland managers with little or no background in computer engineering. It also creates a unified and appealing user interface that farm managers can use. This enables them to analyze complex data intuitively.
\item Robustness/Fault-Tolerance: The system has multiple logging and failure correction mechanisms in place and thus is robust. The failure of one of the UAVs will have little effect on others. Problems like this can be easily solved by replacing hardware that has failed while others continue their normal functions. One obvious point of failure is the ground controller and hardware or software failures. At this point, it can be traced back using logging capabilities that can notify farm owners and the system’s maintainers of newly arisen problems that need to be fixed. This will incrementally make the system less prone to error.
\item Coverage of Large Areas: the system/architecture, through the use of multiple UAVs, can collect data or deliver data on large swaths of land. This might be appealing for farm owners the have huge contiguous areas of land. 
\item Developer Friendly: This architecture is developer-friendly as clearly delineate which parts of the architecture serve what function and thus a developer can jump in and work on a layer that seems suitable to his expertise without breaking the whole system. A developer may, for example, create a new machine learning model for a task and thus needs only to integrate that algorithm into the stack without him being concerned about communication or task allocation.
\end{itemize}
% \subsection{Software Dashboard}
% The dashboard, as shown in the fig was built so that data generated by the UAVs can be integrated and visualized in an orderly manner to end-users who may not have the expertise to run commands to fetch or update information from the system. It also provides a level of abstraction on how the system performs its tasks. 

% \begin{figure}[ht]
%   \includegraphics[width=0.4\textwidth]{screenshot.png}
%   \caption{The Web based user interface for planning mission and monitoring the farm.}
%   \label{our_arc}
% \end{figure}
\subsection{Onboard Computer}
Each UAV will be equipped with onboard computer, to process the image and other data collection. It will serve as a middle man between the Ground Station, and the flight controller. It will be in charge of the following tasks: 
\begin{itemize}
 
\item It will be the requesting and setting parameters for the mission controller, such as the battery level, altitude, GPS data, etc. 
\item It will be tagging the Image and other data collected with GPS position, which will be important to later reconstruct the collected images to visually represent. 
\item It will be in charge of the communication between the Ground station and other member UAVs. 
\begin{itemize}
 
\item It will be notifying the ground station of its current states and battery level and then it will be receiving tasks, waypoints, and the altitude to carry out a mission. After it is done with the mission, it will upload the collected data to the data-offloading unit, clearing a space for the next mission. 
\item In case it can’t carry out a mission, due to malfunction caused on the flight controller or running out of battery power. It will notify the ground station and set up an emergency connection with the other UAVs, to transfer the mission into its neighbor UAVs or cache it for later. 
\end{itemize}
\end{itemize}
\subsection{Data Processing}
RGB and hyperspectral camera's have proven to be useful in many agriculture applications \cite{groupref10A}. However, RGB camera's lacks the spectral range and precision to profile Agricultural farms that only hyperspectral sensors can provide. This kind of high-resolution spectroscopy was first used in satellites and later in manned aircraft, which are significantly expensive platforms and extremely restrictive due to availability limitations and/or complex logistics \cite{groupref10A, groupref10B, groupref10C}. More recently, UAVs have emerged as a very popular and cost-effective remote sensing technology, composed of aerial platforms capable of carrying small-sized and lightweight sensors. 

The advantages of hyperspectral data over RGB imagery is multispectral data are provide not only the Green, Red, Red-Edge but also the Near Infrared wavebands to capture both visible and invisible images of crops and vegetation. The hyperspectral sensors ability to measure hundreds of bands that can be reconstructed later to retrieve different information such as the Normalized Difference Vegetation Index (NDVI) makes them to be widely used. NDVI ratio is a simple graphical indicator of different conditions of the farm such as differentiating bare soil from weed or grass, detecting plants/crops under stress, and differentiate between crops and the crop stages\cite{groupref10D}. 

With recent advances in deep learning, it is also possible to expand RGB imagery to also leverage their functionality into areas that were dominated by spectral cameras. Convolutional Neural Network (CNN ) can be trained to classify, differentiate crops from weed, and can be able to detect plants affected by diseases \cite{groupref10E}. In the architecture, They software framework included a CNN network that is to be able to classify crops and weeds. Test implemented on Soy Beans, managed to get about 96.6\% accuracy on the test set.  Also, the presented architecture is built to be simple for a software developer to expand this into segmentation and detection of different crop types. 
\section{Evaluations}\label{eval}
The presented architecture has been validated by both simulations and real flight tests with UAVs flying simultaneously and working together to gather imagery data, using RGB cameras. Tests are conducted using 2 well known open-source flight stacks, the APM3 and PX4 flight controllers, with onboard computers over them, the Raspberry Pi 3 Model B+,  quad-core, 64-bit processor. A web-based easy to use ground station has been provided where a user can plan a mission, monitor the farmland, and also view the processed image from the collected data that passed through the CNN network, which is presented in a heat-map.  In the following section describes additional evidence related to the presented architecture evaluation on field test and the simulation environment.
\subsection{Simulation Evaluation}
In order to test and evaluate the architecture and software framework,a real-time simulator called Gazebo was used. Gazebo is well-designed simulator which provides a toolbox
to rapidly test algorithms, design robots, perform testing, and train systems using realistic scenarios. With Gazebo it is possible to simulate robots in outdoor environments accurately and efficiently. It is a robust physics engine, high-quality graphics, and convenient programming and graphical interfaces.
\begin{figure}[ht]
\centering
\includegraphics[width=0.4\textwidth]{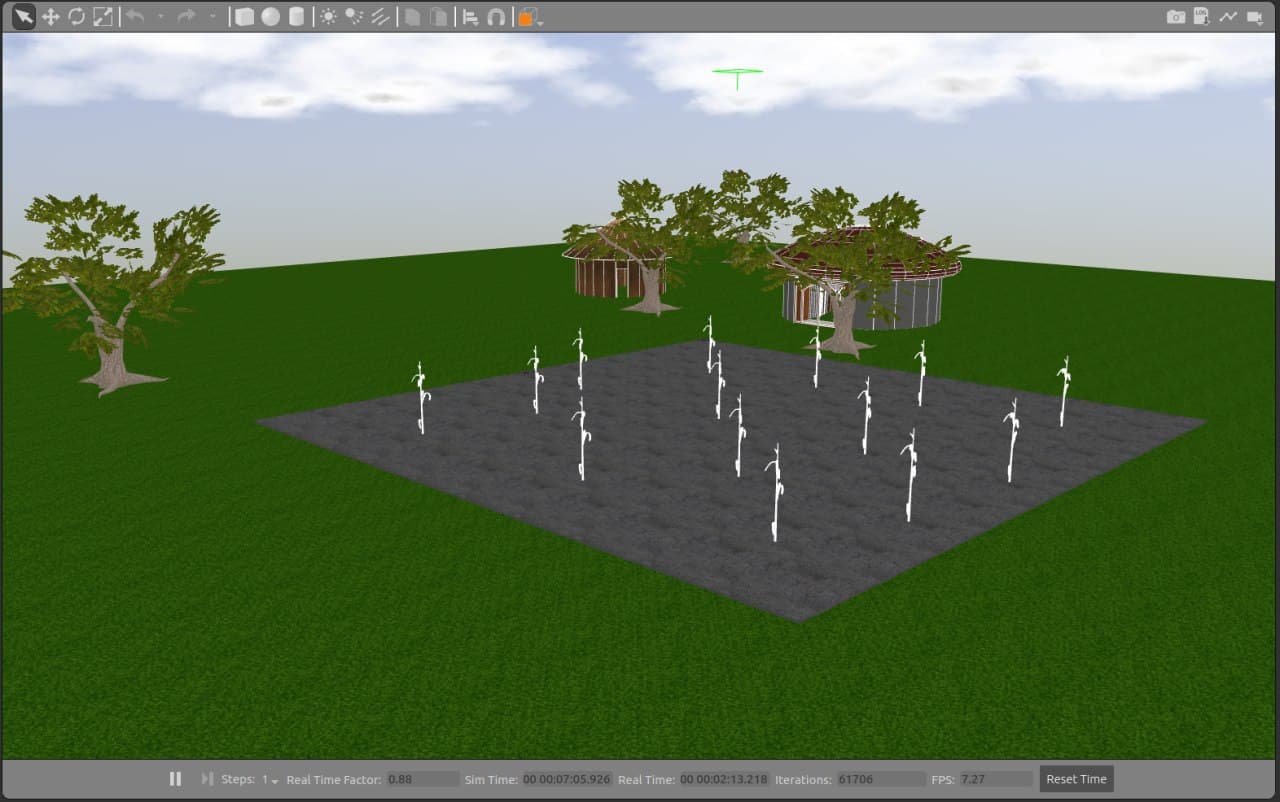}
\caption{Simulation environment created in Gazebo}
\label{farm}
\end{figure}

 In order to test the Multi-UAVs architecture presened in this paper, a custom environment to mimic typical farmland in Ethiopia was created as shown in fig. \ref{farm}, which is about 4 hectares of farmland. 
 
%  the 3D objects from Google’s 3D warehouse and then merged them into our Simulation Description Format (SDF) file to create our environment.

\begin{figure}[ht]
\centering
\includegraphics[width=0.4\textwidth]{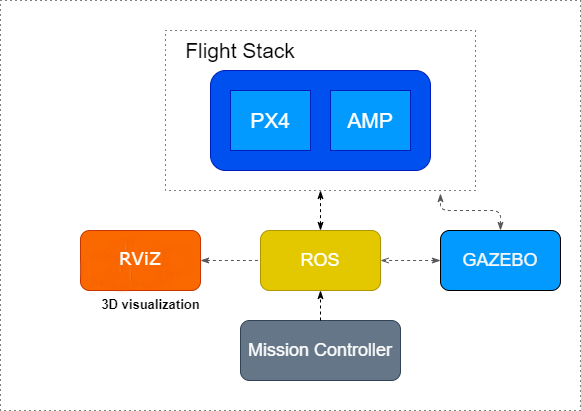}
\caption{Simulation architecture used to test and verify the algorithms and codes. }
\label{software}
\end{figure}

Then the IRIS Quadcopters models, which are provided by the DroneCode PX4 was added. By the firmware provided by the DroneCode PX4 to includes the camera model which streams images via the Ros-topics to Rviz. Rviz (ROS visualization) is a 3D visualizer for displaying sensor data and state information from ROS. Using RViZ, its possible to visualize images from the UAVs in the gazebo simulator. In order to test the algorithms provided by the software architecture, a modified version of the OpenAI Gym was used \cite{groupref11} as shown in the fig. \ref{software}.

% Once tested on had our models and environment set up, we then wrote a python script to take control of the motion of the Drone in the environment. This script allows our drone to navigate over the region. The script uses data registered through the dashboard using the same process that we would have used in a hardware environment. Thus, this simulation can be transferable to real-life runs. The system uses the same mission planning, task allocation, and other functions used in the main system. 
\subsection{Hardware Evaluation}
To demonstrate the operation of the proposed architecture in the field, a flight tests were conducted in Addis Ababa University Campus to classify grass and soil.

For this purpose, two UAVs with the PX4 and APM3 were used as flight controllers, both mounted with an onboard computer like the Raspberry Pi3 to collect and store data temporarily. Using the web-based user interface provided, a selected region of interest which was a football field in the campus was provided for the UAVs. Once the mission is generated, and UAVs are connected automatically, a JSON file containing a region, waypoints, and travel instructions was sent to the onboard computer on the UAVs. With the serial communication, it was possible to get the GPS location of the UAVs during the flight from the flight stacks. This was then used to tag the image that the Raspberry Pi was collecting. This image was finally uploaded to the server for Image analysis. 
\begin{figure}[ht]
\centering
\includegraphics[width=0.4\textwidth]{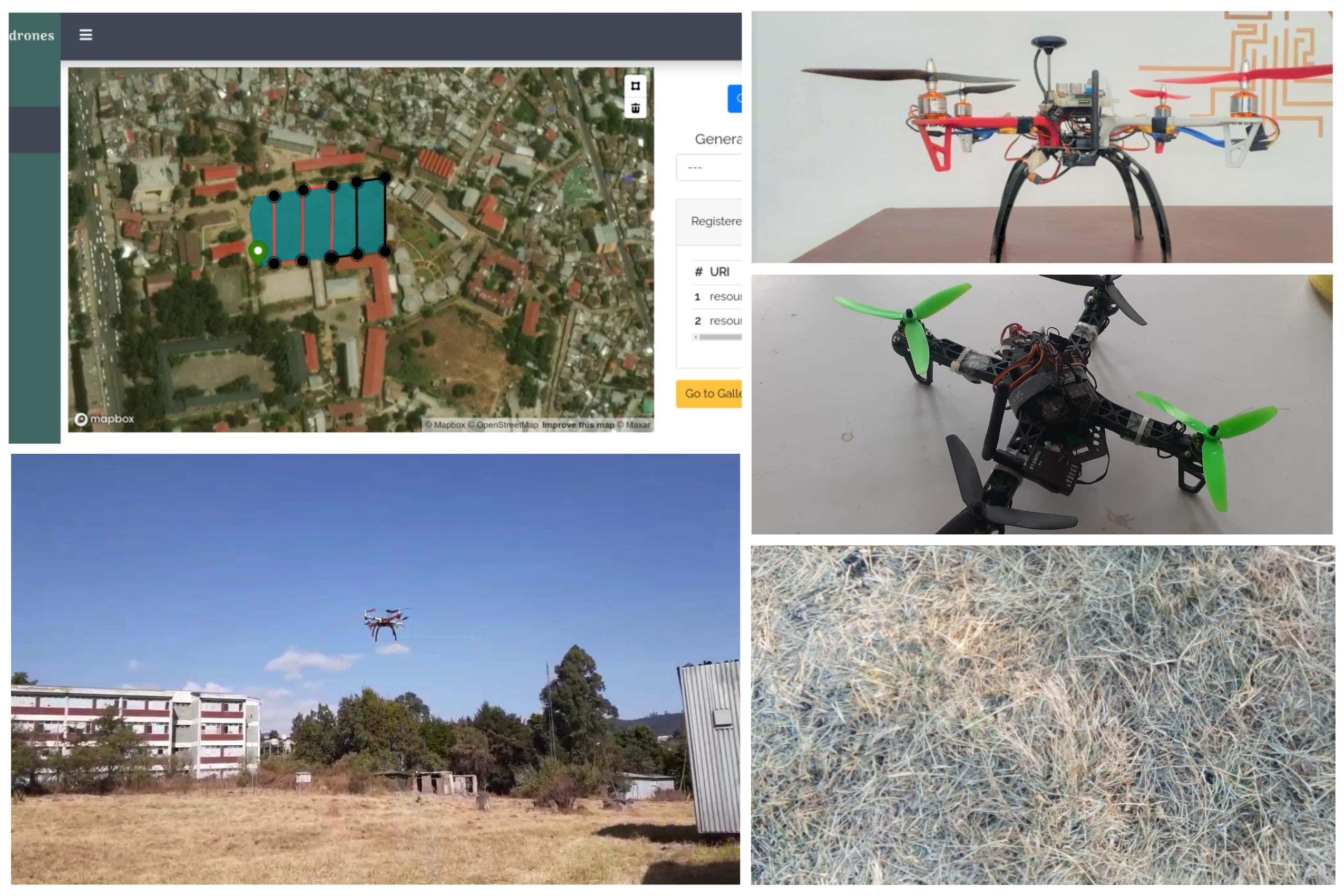}
\caption{The test set-up used, starting with the dashboard to load a mission, the two quad-copters used and the sample image they collected}
\end{figure}
\subsection{Performance of Deep Learning Model}
The proposed architecture used a well-known image classification algorithm known as CNN, as most Intelligent spot-spraying systems and crop monitoring predominantly rely on machine vision-based detectors for autonomous weed control \cite{groupref12}. The soybean crop data-set  from  Kaggle  and  trained was used to train the CNN model on the Google Colab platform, since it offers a faster training hour model using Google’s faster GPU.
\begin{figure}[ht]
\centering
\includegraphics[width=0.45\textwidth]{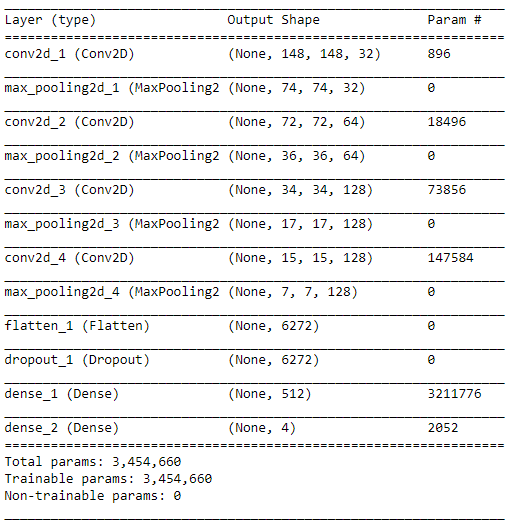}
\caption{The CNN model used to classify the collected image datas.}
\end{figure}

The data set for training is captured by the UAV, all those with the occurrence of weeds were selected resulting in a total of 400 images. Through the Pynovisão software, using the SLIC algorithm, these images were segmented and the segments annotated manually with their respective class \cite{groupref12a}. These segments were used in the construction of the image dataset. This image dataset has 15336 segments, being 3249 of soil, 7376 of soybean, 3520 grass and 1191 of broadleaf weeds. 

The neural network summary is provided in the figure above. The model is set up of Four 2D Convolution layers, each followed by a max-pooling layer, which then passed through a Fully connected (FC) layer, which finally passed through a softmax activation function to provide a predicted probability of the class. 

% The probability is used to reconstruct the heatmap that will be displayed to the user to show areas affected by the weed and grass from the crop, fig. \ref{heatmap}.

After training for 15 epochs, it was possible to get a test accuracy of about 96.6\% as shown in the graph below. 
\begin{figure}[ht]
\centering
\includegraphics[width=0.5\textwidth]{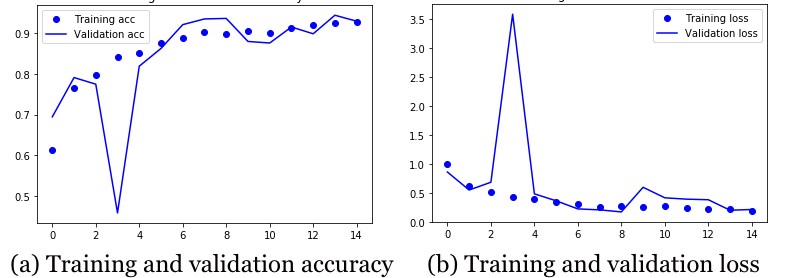}
\caption{The performance of the CNN network, showing the accuracy on fig. (a) increasing with each epoch and in fig. (b) the loss decreasing with the epochs. }
\label{perfo}
\end{figure}

\section{Conclusion and Future Work}\label{conc}
The adoption of UAVs in the agriculture sector has already established paradigms to increase farm productivity and quality, as well as improving working conditions. Even though the existing frameworks and fight stacks have demonstrated a good level of autonomous operation in different applications, they don’t offer a complete end-to-end solution in agriculture. 

In this paper, Multi-UAVs architecture was presented to provides a multi-UAV framework integrating different elements and algorithms that have been tested and proven to be robust and developer friendly.  The software framework also enables distribution and scalability as well as extensibility of components.  Additionally, the elements such as BSS will optimize the power efficiencies and give the multi-UAV system more autonomy and increase the Fault-Tolerance issues that might have arisen due to the power efficiency of the system. Besides, when the farmland increases having BSS elements set-up will be vital in collecting and having data collected with efficient power usage that would have otherwise been spent on their return to the charging station. 

The presented architecture has been tested by on the software framework provided,  starting with a web-based ground station which is used to set up a region of interest for the UAVs to collect data.  From this selected ROI, waypoints are optimized to have less number of turns and the path for each member UAVs will be generated.  Then the generated paths are sent to the Onboard computer(Raspberry Pi) which is found on the UAVs. The onboard computer was the one in charge of the flight control and collecting, storing them during its flight time. As specified in the Architecture it has to make a decision such as to battery issues, for instants. After collecting and finally uploading the datas to the Deep learning algorithms, the CNN classified the images in the mapped region on the ground station. Once that process is done, it will present a heat map of the region to show the areas affected with either weed or diseases. In the Future, the Architecture will include a reactive UAVs such as, a pesticide sprayers in the spotted region to combat the plant diseases and weeds.

There are several potential points to improve the proposed architecture. For one, task allocation can be improved by adding additional variables to the problem. Although this makes the complexity of the algorithm higher and would lead to an algorithm solving optimization problems, it will definitely help to create a more efficient solution to the problem. Another point to work on is decentralization and ad hoc message passing when controllers fail at their task. In these cases, its possible to propose multiple solutions like assigning a temporary master using election algorithms or enable swarm-like tendencies for the UAVs. Overall, the system is built on easily accessible technologies, and when, over time, communication technologies increase in capability, the architecture will be integrating those that are applicable.

% or


        % \appendices
        % \section{}
        % Do we really need one?

 \section*{Acknowledgment}
 
 The authors would like to thank, the Computer Science department in Addis Ababa University, especially the 4k Lab, for providing space and materials. 
%  % <-this % stops a space

\ifCLASSOPTIONcaptionsoff
  \newpage
\fi

\bibliographystyle{unsrt}
\bibliography{refs}

\end{document}